\documentclass{article}
\usepackage{spconf,amsmath,epsfig,enumerate,xcolor,fancyhdr,hyperref}

\newcommand{\enumc}{\begin{enumerate}[$\circ$]}
\newcommand{\enum}{\begin{enumerate}[i)]}
\newcommand{\eenum}{\end{enumerate}}

\title{Bridging the Gap: Simultaneous Fine Tuning for Data Re-Balancing}
\name{John McKay$^{1, 2}$, Isaac Gerg$^2$, \& Vishal Monga$^1$\thanks{Funding provided by ONR N00014-15-1-2042}\thanks{Acknowledgements to Tiantong Guo$^1$, Tiep Vu Huu$^1$ for code advice}}
\address{Dept of Electrical Engineering \& Computer Science$^1$, Applied Research Laboratory$^2$\\ Pennsylvania State University}

\begin{document}
%
\maketitle
\begin{abstract}
There are many real-world classification problems wherein the issue of data imbalance (the case when a data set contains substantially more samples for one/many classes than the rest) is unavoidable. While under-sampling the problematic classes is a common solution, this is not a compelling option when the large data class is itself diverse and/or the limited data class is especially small. We suggest a strategy based on recent work concerning limited data problems which utilizes a supplemental set of images with similar properties to the limited data class to aid in the training of a neural network. We show results for our model against other typical methods on a real-world synthetic aperture sonar data set. Code can be found at \href{putCodeHere}{\color{blue}\texttt{github.com/JohnMcKay/dataImbalance}}.
\end{abstract}
\section{Introduction}
\label{sec:intro}
The goal of any ``re-balancing" scheme is to convince an algorithm to not disregard an underrepresented class. This is nontrivial as learned algorithms are incentivized to perform well on their training and if they see an overwhelming number of a certain class, they are going to be more apt to classify inputs in that direction. When it comes to sonar target recognition, it is obvious that such a tendency will be dangerous.

How do people typically handle data imbalances? For natural, optical images problems it is common to under-sample the large class \cite{he2009learning}. This means purposefully removing training samples when training, say, a neural network. This and variants that compensate using synthesized data may work well in certain cases \cite{barua2014mwmote}, but when it comes to sonar (or radar), omitting background elements that could contain unique debris or distinctive rocky patches is heading towards a direction of \emph{less} information making it to the model. This can lead to confusion later on when tested on field data.

In \cite{ge2017borrowing}, the authors present a novel manner of training convolutional neural networks (CNNs) when dealing with small data sets. They suggest drawing images from a supplemental, larger (source) data set that shares low-level features with the target data and, when training a CNN, propose simultaneously learning shared initial layers with the target and source data sets. This helps alleviate over-fitting and allows the deeper, finer layers to extract more information. We see that such a strategy can be adapted for the data imbalance problem in that a supplemental collection of images from a source data set can be used to draw images that are similar with respect to low level features to the data limited class and are dissimilar to the larger class. This discriminative parsing of the supplemental data set corrects for the data imbalance while not sacrificing information pertaining to the larger class.

In the following, we look to: formulate and detail our novel discriminative adaptation of \cite{ge2017borrowing} for data imbalances and demonstrate its potential with an undersea identification problem using real synthetic aperture sonar (SAS) images. Section \ref{sec:features} goes through how images from a supplemental data set are chosen. Section \ref{sec:model} goes through our CNN architecture and how the simultaneous training works. Lastly, Section \ref{sec:experiments} contains experimental work using the aforementioned actual SAS data. Note, we use \emph{target data set ($D_t$)} to refer to the entire collection of images we want to classify, \emph{source data set ($D_s$)} to refer the set of images we are drawing from to supplement our target data set, \emph{starved class ($c_s$)} to refer to the limited data class of $D_t$, and \emph{large class ($c_\ell$)} to refer to the larger data class of $D_t$.

\section{Supplemental Data Selection}
\label{sec:features}

\begin{figure*}[t]
\centering
\fbox{
\begin{minipage}{.31\textwidth}\centering
\includegraphics[width=\columnwidth]{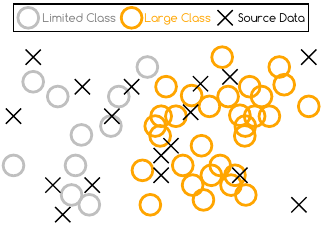}
\caption{Example feature space of both classes of $D_s$ and members of $D_s$.}\label{fig:histogramSpace}
\end{minipage}}
\fbox{
\begin{minipage}{.31\textwidth}\centering
\includegraphics[width=\columnwidth]{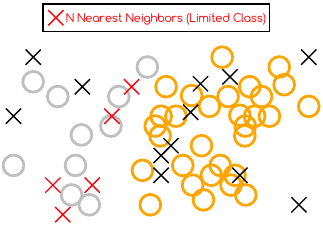}
\caption{Feature distances calculated and $N$ closest to $c_s$ selected ({\color{red}red}).}\label{fig:K1}
\end{minipage}}
\fbox{
\begin{minipage}{.305\textwidth}\centering
\includegraphics[width=\columnwidth]{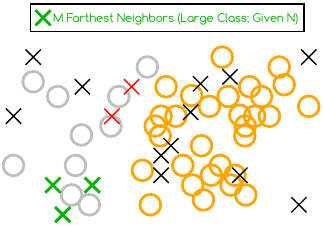}
\caption{Given $N$ from Fig. \ref{fig:K1}, $M$ farthest from $c_\ell$ kept ({\color{green}green}).}\label{fig:K2}
\end{minipage}}
\end{figure*}

Our scheme for ``re-balancing'' data depends highly on picking the ``right'' images from $D_s$. That is, we need to select images from $D_s$ that are similar to $c_s$ and dissimilar to $c_\ell$ all according to some metric that detects meaningful characteristics. Let's start with the metric aspect; we know that trained convolutional neural networks have initial layers that resemble Gabor filters \cite{yosinski2015understanding}. These edge filters are then followed in networks by layers that extract finer and finer details \cite{oquab2014learning}. For our purposes, the later layers are not of much interest; we do not expect images in, say, Imagenet \cite{imagenet} to share high level characteristics with lobster crates under the sea. That said, we can expect that some images to share similar low-level edge characteristics. This is the crux of the idea behind using histograms of Gabor filter activations as the feature vectors for deciding ``nearness.'' For every image in $D_t$ and $D_s$ we use Gabor filters to get edge maps and then get histograms of their intensity values. These activation histograms are then concatenated according to their image and this vector serves as our feature transformation. 

This is analogous in some ways to what \cite{ge2017borrowing} did with limited data. They used the filters of an existing, trained network (Alexnet \cite{krizhevsky2012imagenet}) in conjunction with Gabor filters. These additional filters were only marginally helpful for our purposes and not worth the larger vector and increased computational stress. When dealing with more intricate images than our SAS example (i.e. ones that have color, have higher resolution, etc) such an action may be warranted.

With these feature vectors, the next step is to decipher which are close to $c_s$ and far from $c_\ell$ to which we employ a nearest/farthest neighbor approach. We go through the edge features according to every source image and find their distance to the members of $D_t$.  We then parse this data in two steps: first pluck out the $N$ closest source images to a member of $c_s$ and then refine that list to the $M<N$ farthest from the members of $c_\ell$.  The idea is that we first want to ensure that our pool of potential supplemental images are, foremost, similar to the limited class and then we can make selections based on the distances away from the large class. This two stage approach does not require any additional calculations beyond a single nearest neighbor implementation, but this is still a nontrivial computational expense. We suggest using a random sample $\hat D_s\subset D_s$ to keep computations manageable. 

Overall, the idea is this: for each $y\in D_t$, obtain their histogram features according to filter $f^{(i)}$, $h^{(i)}_y$ and concatenate each one over $i=1,\dots, F$ histograms into a single vector $h_y=[h^{(1)}_y,\dots,h^{(2)}_y]^T$. Do the same for the images in the source data set, $x\in D_x$, to obtain $h_x$ and then calculate the distance between each of the source and target histograms, arriving in a scenario shown in Figure \ref{fig:histogramSpace} (We suggest the L2 norm). The final selection is then done in two parts: isolate the $N$ source images that have the smallest distance between them and a member of $c_s$ and then, of that set, sort them by their nearest distances to a member of $c_\ell$ and keep the $M$ farthest.  Both parts of the final step are described in Figures \ref{fig:K1} and \ref{fig:K2}. We let $U\subset \hat D_s$ be the set of images from $\hat D_s$ that have been chosen to supplement class $c_s$. 

Note that, in some ways, we are designing a scheme similar to the well-known SMOTE method used for support vector machine and similar classifiers \cite{chawla2002smote,akbani2004applying}. Instead of crafting synthetic samples, our use of existing data circumvents a generation step. This means we in principal are pursuing the same idea as SMOTE (obtaining representative features for learning) with a clever work around tailored for neural networks.

\section{Simultaneously Trained Network}
\label{sec:model}

The reason we get $U$ is so that we can use it for training. Before we go forward, it is worth mentioning the common practice of weight sharing and transfer learning. When dealing with limited data or initialization problems, there is an idea of using existing weights from heavily trained models like VGGnet \cite{simonyan2014very}, Alexnet, etc. where authors had ample resources to train their networks on millions of images. Since the differences between natural images are relatively small, models can be \emph{finely tuned} by starting with those existing weights and then trained from there with the target data set \cite{oquab2014learning}. 

Simultaneously trained models (STMs) are similar in concept but differ in implementation. STMs start from randomized weights and flip between batches of $U$ and $D_s$. This means, instead of crafting a model using the source data set and then imposing new information via fine tuning with $D_t$, STMs start anew and have $D_t$ and $U$ struggle against one another during the entirety of training. As the only shared quality between $D_t$ and $U$ are the edge features, this keeps STMs from getting overly influenced by $c_\ell$ regardless of how long they are trained. This is crucial for our re-balancing scheme.

\begin{figure*}\centering
\fbox{
\begin{minipage}{.75\textwidth}
\includegraphics[width=\columnwidth]{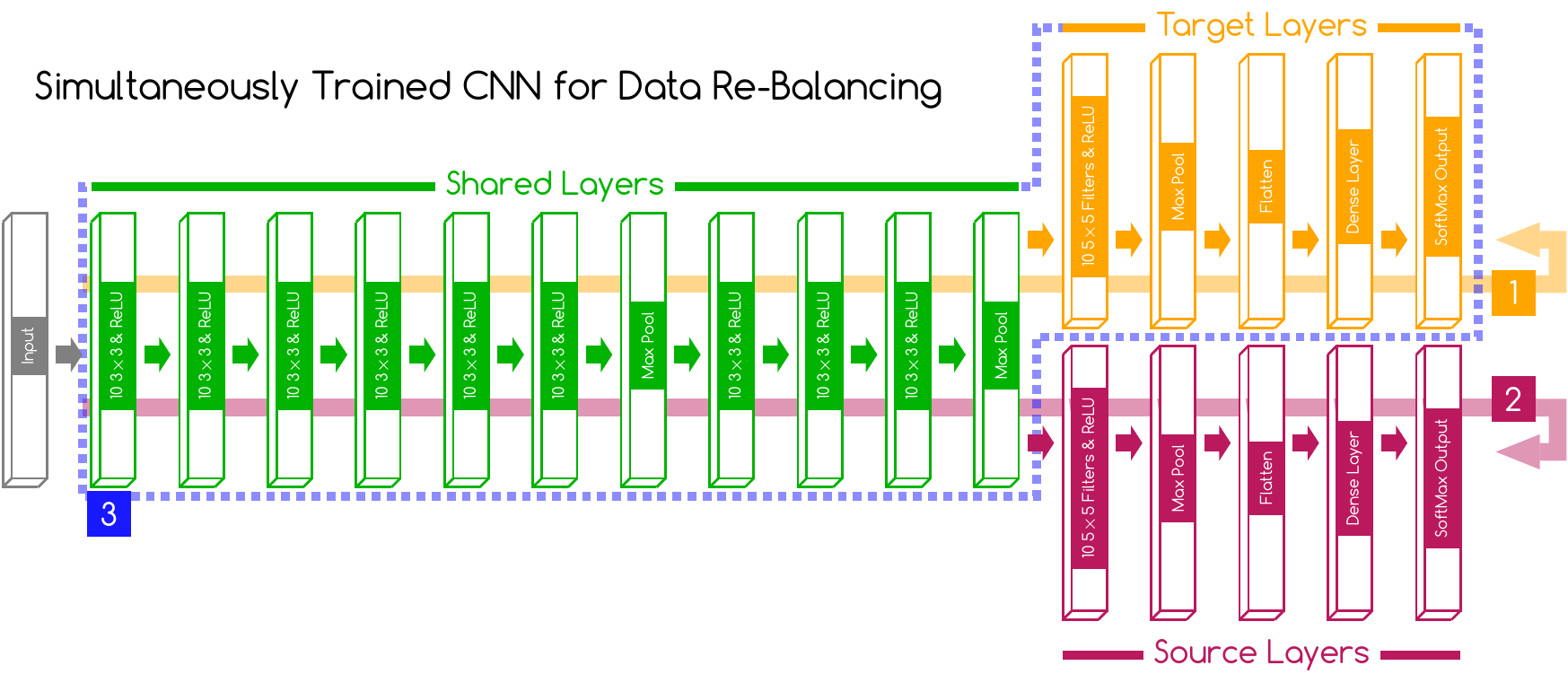}
\caption{A diagram of our simultaneous finely tuned CNN for imbalanced data. Training goes as follows: ``1.5'' models are made (one set of shared layers that feed into designated target layers and designated source layers) and a batch from $D_t$ is fed through (1) with weight updated followed by a (smaller) batch from $D_s$ through (2) and, again, weights updated. After sufficiently many epochs, the source layers are dropped and (3) is used for testing. The given architecture is what we used for section \ref{sec:experiments}.}\label{fig:model}
\end{minipage}}
\fbox{
\begin{minipage}{.2\textwidth}\centering
\includegraphics[width=.48\columnwidth]{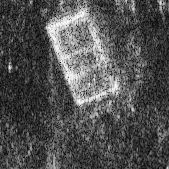}
\includegraphics[width=.48\columnwidth]{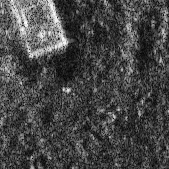}\\\vspace{.03cm}
\includegraphics[width=.48\columnwidth]{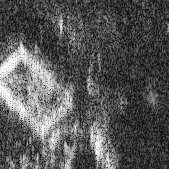}
\includegraphics[width=.48\columnwidth]{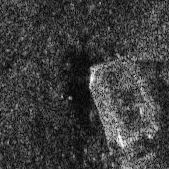}\\\vspace{.03cm}
\includegraphics[width=.48\columnwidth]{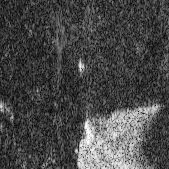}
\includegraphics[width=.48\columnwidth]{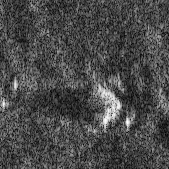}\\\vspace{.03cm}
\includegraphics[width=.48\columnwidth]{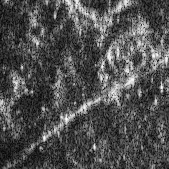}
\includegraphics[width=.48\columnwidth]{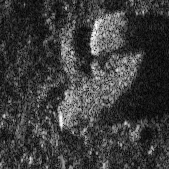}
\caption{Lobster crates (top four) and clutter (bottom four) images.}\label{fig:noaa}
\end{minipage}}
\end{figure*}

\section{Sonar Target Recognition}
\label{sec:experiments}
Sonar automatic target recognition (ATR) suffers from extreme data imbalances \cite{stack2011automation} and we designed an experiment to illustrate the potential of our method as a viable option in this domain. We looked to classify objects (lobster crates) from undersea clutter using real-world data. Our images came from a synthetic aperture sonar system equipped to an unmanned underwater vehicle that scoured along New England's coast. The entire data set consisted of approximately 1.71km$^2$ of underwater area coverage. 1.12km$^2$ was designated as training and supplied 169,413  168$\times$168 patches with 869 of those containing crates (which we upsampled by a factor of ten for data augmentation) and the rest clutter (a 194:1 background-to-crate ratio).  0.58km$^2$ was used as the testing set and gave 89,048 patches consisting of 757 with crates and the other 88,291 as clutters (a 117:1 background-to-crate ratio).

Evaluating the effectiveness of a classifier on imbalanced data is nuanced. Typically, one sees a ROC curve when dealing with a binary problem but they are ill-equipped for skewed data scenarios; ROC curves \emph{under emphasize} the effect of large numbers of false positives \cite{davis2006relationship}. Instead, we looked at two more informative metrics: precision-recall (PR) and false alarm rate (FAR) curves. PR curves are less biased than ROC curves as they do not consider true negatives which overwhelm ROC curve statistics \cite{davis2006relationship}. FAR curves replace a ROC curve's false positive rate with the expected number of false alarms per square 0-1km (i.e. a hard cut-off at 1km$^2$) and are standard for ATR problems \cite{ross1998standard}.

\begin{figure*}[t]\centering
\fbox{
\begin{minipage}{.31\textwidth}
\includegraphics[width=\columnwidth]{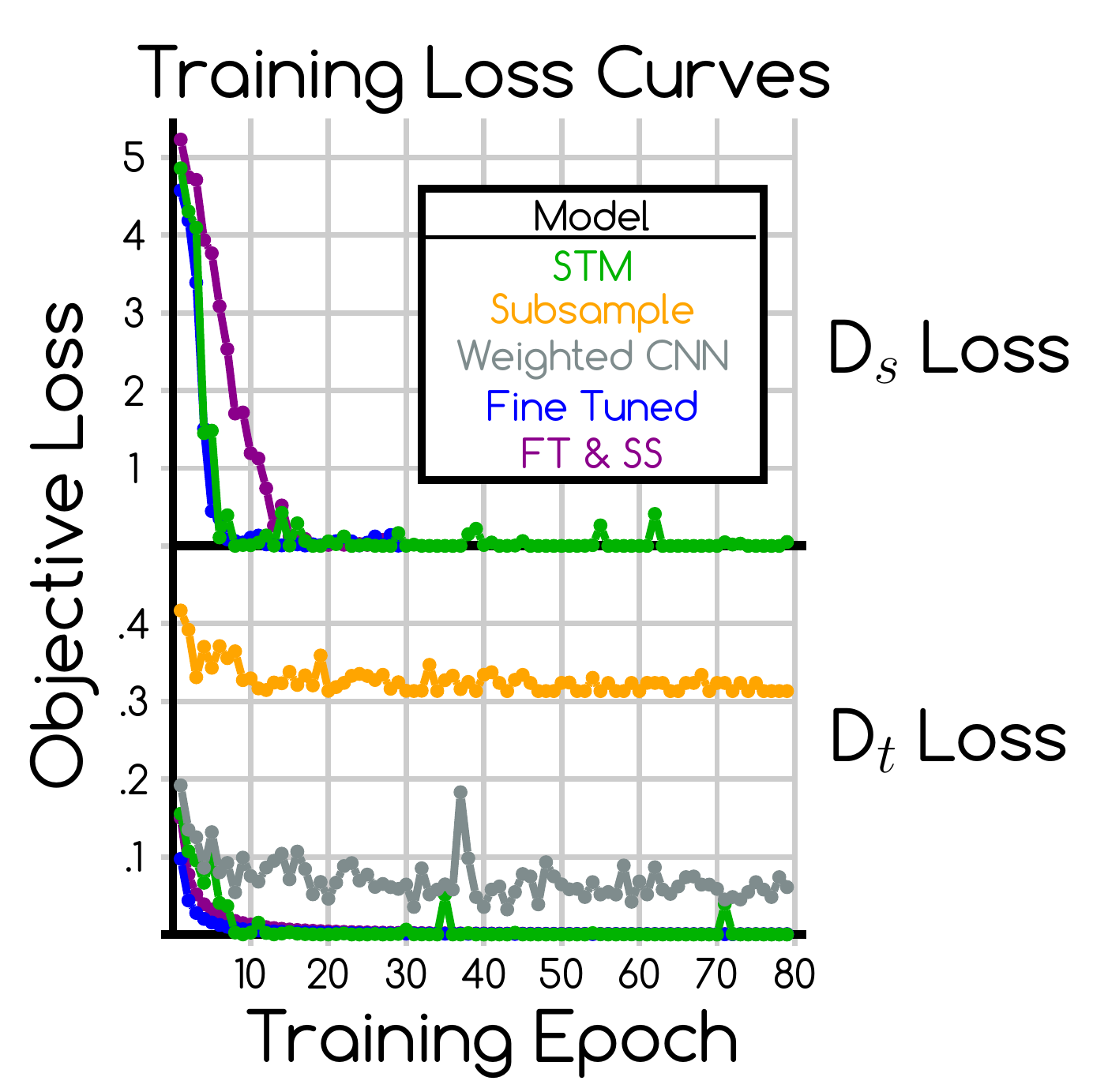}
\caption{Per-epoch training function loss. Top is $D_s$ or $U$ loss, bottom is $D_t$. Training parameters found in \href{putCodeHere}{\color{blue}\texttt{code}}.}\label{fig:loss_curves}
\end{minipage}}
\fbox{
\begin{minipage}{.31\textwidth}
\includegraphics[width=\columnwidth]{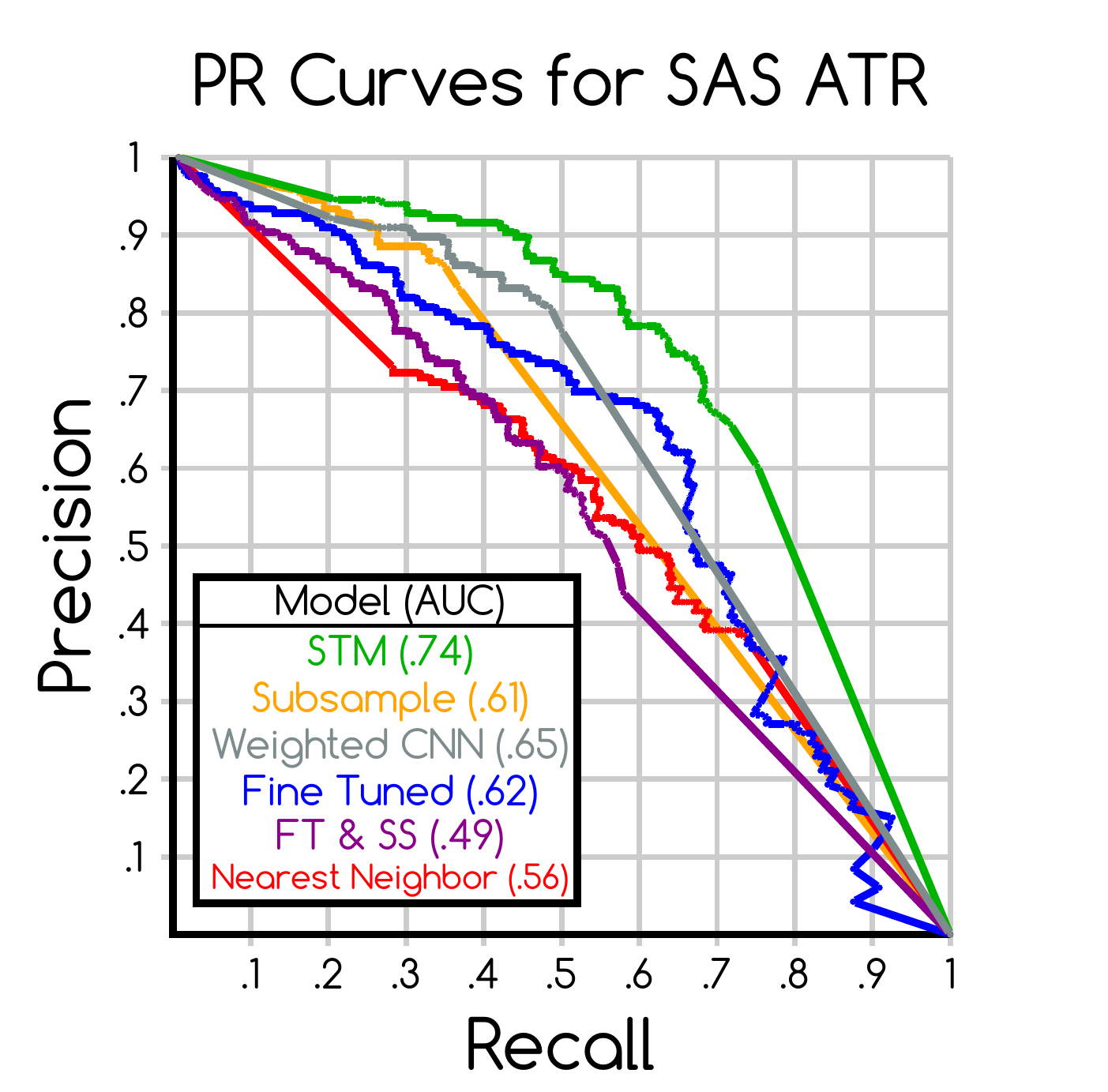}
\caption{Precision-Recall for five CNN models and nearest neighbor. AUC is the mean precision over all tests.}\label{fig:pr_curves}
\end{minipage}}
\fbox{
\begin{minipage}{.31\textwidth}
\includegraphics[width=\columnwidth]{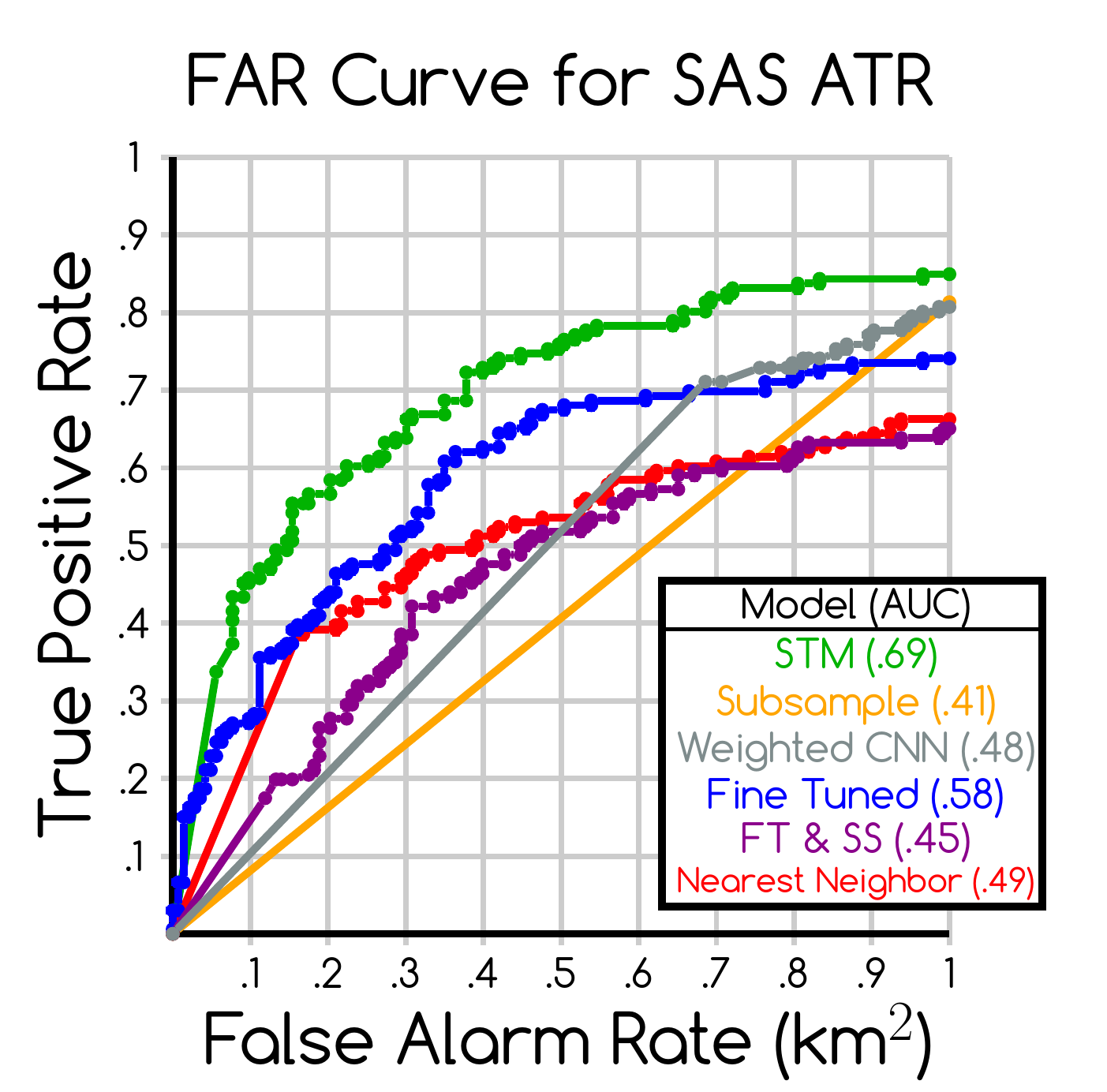}
\caption{False alarm rate per km$^2$ for five CNN models and nearest neighbor. AUC is the discrimination over 1 km$^2$.}\label{fig:far_curves}
\end{minipage}\vspace{.06cm}}
\end{figure*}

We looked to use the Caltech 256 data set \cite{griffin2007caltech} as a $D_s$ and built a STM with $N$=18,000 and $M$=12,000. For context, we also built: a CNN with no source data influence and trained with a subsampling of the larger class, a CNN with class weights trained on an imbalanced data set, a CNN that was fine tuned using $D_s$ and then trained on the imbalanced set, a CNN that was fine tuned with $D_s$ and then trained using a subsampled data set, and, lastly, a simple nearest neighbor (i.e. non-CNN) scheme that used a distance-based scoring metric so we could illustrate its PR/FAR performance. Each competing scheme offers a different view of against our STM model. Note for the CNNs, we used evaluated the AUC of a PR curve on a small validation set at each epoch and chose the weights based on the optimal epoch. The fine tuned models were pre-trained on $D_s$ for 30 epochs based on over-fitting and general performance.

In our experiments, the STM outperformed the five others in testing and, as Figures \ref{fig:pr_curves} and \ref{fig:far_curves} show, there was a considerable gap. Our results revealed a trend: models that could use the full, imbalanced training did better than subsampled ones. Even the nearest neighbor scheme with the full training set did better in terms of FAR than the two subsampled cases, reflecting the benefit of more information. The STM's relative success unveils the power in our discriminative, simultaneous training scheme; if $D_s$ were to be used for fine tuning without any refinement, it is arguably as powerful as just weights. 

We lastly note an interesting phenomenon with regards to the training loss. As shown in Figure \ref{fig:loss_curves}, the $D_s$-using models asymptoted to zero yet the fine tuned with subsampling failed to achieve the quality of the others, suggesting a different local minima convergence. In later work, it would be prudent to investigate the loss function manifold to understand how our and other training schemes impact its geometry.

\section{Conclusion}
\label{sec:conclusion}
We have shown that STMs using a discriminatively chosen source set $U$ can help alleviate the problematic trade-off between incorporating more large-class information into a model and the bias that causes. A further investigation into the choice of $D_s$ or edge-feature statistics may be a fruitful direction for future research, but for now we consider our work a compelling option for those struggling with imbalanced training sets, especially in the case of sonar ATR.

\bibliographystyle{IEEEbib}
\bibliography{refIGARSS2018}

\begin{thebibliography}{10}

\bibitem{he2009learning}
H.~He and E.~A. Garcia,
\newblock ``Learning from imbalanced data,''
\newblock {\em IEEE Transactions on Knowledge and Data Engineering}, vol. 21,
  no. 9, pp. 1263--1284, Sept 2009.

\bibitem{barua2014mwmote}
S.~Barua, M.~M. Islam, X.~Yao, and K.~Murase,
\newblock ``Mwmote--majority weighted minority oversampling technique for
  imbalanced data set learning,''
\newblock {\em IEEE Transactions on Knowledge and Data Engineering}, vol. 26,
  no. 2, pp. 405--425, Feb 2014.

\bibitem{ge2017borrowing}
Weifeng Ge and Yizhou Yu,
\newblock ``Borrowing treasures from the wealthy: Deep transfer learning
  through selective joint fine-tuning,''
\newblock in {\em The IEEE Conference on Computer Vision and Pattern
  Recognition (CVPR)}, July 2017.

\bibitem{yosinski2015understanding}
Jason Yosinski, Jeff Clune, Anh Nguyen, Thomas Fuchs, and Hod Lipson,
\newblock ``Understanding neural networks through deep visualization,''
\newblock {\em arXiv preprint arXiv:1506.06579}, 2015.

\bibitem{oquab2014learning}
Maxime Oquab, Leon Bottou, Ivan Laptev, and Josef Sivic,
\newblock ``Learning and transferring mid-level image representations using
  convolutional neural networks,''
\newblock in {\em Proceedings of the IEEE conference on computer vision and
  pattern recognition}, 2014, pp. 1717--1724.

\bibitem{imagenet}
J.~Deng, W.~Dong, R.~Socher, L.-J. Li, K.~Li, and L.~Fei-Fei,
\newblock ``{ImageNet: A Large-Scale Hierarchical Image Database},''
\newblock in {\em CVPR09}, 2009.

\bibitem{krizhevsky2012imagenet}
Alex Krizhevsky, Ilya Sutskever, and Geoffrey~E Hinton,
\newblock ``Imagenet classification with deep convolutional neural networks,''
\newblock in {\em Advances in neural information processing systems}, 2012, pp.
  1097--1105.

\bibitem{chawla2002smote}
Nitesh~V Chawla, Kevin~W Bowyer, Lawrence~O Hall, and W~Philip Kegelmeyer,
\newblock ``Smote: synthetic minority over-sampling technique,''
\newblock {\em Journal of artificial intelligence research}, vol. 16, pp.
  321--357, 2002.

\bibitem{akbani2004applying}
Rehan Akbani, Stephen Kwek, and Nathalie Japkowicz,
\newblock ``Applying support vector machines to imbalanced datasets,''
\newblock {\em Machine learning: ECML 2004}, pp. 39--50, 2004.

\bibitem{simonyan2014very}
Karen Simonyan and Andrew Zisserman,
\newblock ``Very deep convolutional networks for large-scale image
  recognition,''
\newblock {\em arXiv preprint arXiv:1409.1556}, 2014.

\bibitem{stack2011automation}
Jason Stack,
\newblock ``Automation for underwater mine recognition: current trends and
  future strategy,''
\newblock in {\em SPIE Defense, Security, and Sensing}. International Society
  for Optics and Photonics, 2011, pp. 80170K--80170K.

\bibitem{davis2006relationship}
Jesse Davis and Mark Goadrich,
\newblock ``The relationship between precision-recall and roc curves,''
\newblock in {\em Proceedings of the 23rd international conference on Machine
  learning}. ACM, 2006, pp. 233--240.

\bibitem{ross1998standard}
Timothy~D Ross, Steven~W Worrell, Vincent~J Velten, John~C Mossing, and
  Michael~Lee Bryant,
\newblock ``Standard sar atr evaluation experiments using the mstar public
  release data set,''
\newblock in {\em Algorithms for Synthetic Aperture Radar Imagery V}.
  International Society for Optics and Photonics, 1998, vol. 3370, pp.
  566--574.

\bibitem{griffin2007caltech}
Gregory Griffin, Alex Holub, and Pietro Perona,
\newblock ``Caltech-256 object category dataset,''
\newblock 2007.

\end{thebibliography}

\end{document}